\documentclass[pdflatex,sn-mathphys]{sn-jnl}

\jyear{2021}%

\theoremstyle{thmstyleone}%
\theoremstyle{thmstyletwo}%

\theoremstyle{thmstylethree}%

\begin{document}

\title[Do GPT Language Models Suffer From Split Personality Disorder?]{Do GPT Language Models Suffer From Split Personality Disorder? The Advent Of Substrate-Free Psychometrics}


\author*[1,2]{\fnm{Peter} \sur{Romero}}\email{rp@keio.jp}
\author[3]{\fnm{Stephen} \sur{Fitz}}\email{stephenf@keio.jp}
\author[1]{\fnm{Teruo} \sur{Nakatsuma}}\email{nakatuma@econ.keio.ac.jp}

\affil*[1]{\orgdiv{Faculty of Economics}, \orgname{Keio University}, \orgaddress{\street{2-15-45 Mita}, \city{Minato-ku}, \postcode{108-8345}, \state{Tokyo}, \country{Japan}}}

\affil*[2]{\orgdiv{The Psychometrics Centre}, \orgname{University of Cambridge}, \orgaddress{\street{Trumpington Street}, \city{Cambridge}, \postcode{CB2 1AG}, \state{Cambridgeshire}, \country{United Kingdom}}}

\affil*[3]{\orgdiv{Faculty of Science and Technology}, \orgname{Keio University}, \orgaddress{\street{3-14-1 Hiyoshi}, \city{Kohoku-ku}, \postcode{223-8522}, \state{Kanagawa}, \country{Japan}}}

\abstract{
  Previous research on emergence in large language models shows these display apparent human-like abilities and psychological latent traits.
  However, results are partly contradicting in expression and magnitude of these latent traits, yet agree on the worrisome tendencies to score high on the Dark Triad of narcissism, psychopathy, and Machiavellianism, which, together with a track record of derailments, demands more rigorous research on safety of these models.
  We provided a state of the art language model with the same personality questionnaire in nine languages, and performed Bayesian analysis of Gaussian Mixture Model, finding evidence for a deeper-rooted issue.
  Our results suggest both interlingual and intralingual instabilities, which indicate that current language models do not develop a consistent core personality. 
  This can lead to unsafe behaviour of artificial intelligence systems that are based on these foundation models, and are increasingly integrated in human life.
  We subsequently discuss the shortcomings of modern psychometrics, abstract it, and provide a framework for its species-neutral, substrate-free formulation.
  }




\maketitle

\section{Introduction}\label{sec1}

In Stanley Kubrick's 1968 classical science fiction movie ``2001: A Space Odyssey'', an artificial intelligence, ``HAL'', goes berserk, which unfortunately also runs their spacecraft and all life-support-systems during a mysterious mission to Jupiter.
The name ``HAL'' happens to be a one-letter-shift of IBM, the company spearheading with its Watson division the field of consumer-facing and decision making artificial intelligence.
Though originally based on so called ``good old fashioned AI'', a synonym for rule-based or logical agents, the precursor of nowadays's neural architectures, it won in 2011 against human players in Jeopardy \cite{ferrucci2012introduction}, and was subsequently updated and deployed in various fields from cooking, to code creation, weather forecasting, advertisement, finance, fashion, defence, education, and general chatbots.
One remarkable application was its now deprecated service for deriving author personality from text, IBM Watson Personality Insights, which was mainly geared towards marketing clients and trained on data from people who took personality questionnaires and provided text samples.
The notion of machine personality inspired not only countless science fiction authors and researchers.
Google AI's chatbot ``LaMDA'' was described as `sentient' by Blake Lemoine, a researcher working with it, which became a global news story.
Conversational AI had it's watershed moment however, as ``ChatGPT'', or GPT 3.5 appeared \textit{deus ex machina} and over night influenced culture world-wide.

Given the trend in the industry to intermingle AI with human life spaces through self-driving cars, neural interfaces, ambient artificial assistants, and decision making algorithms, a variety of researchers applied psychometric instruments that were created for humans towards Large Language Models (LLMs). 
These approaches and findings can be clustered into two major categories: emergent latent psychological traits, and emergent abilities.


In terms of emergent abilities, ChatGPT displays human-like ability to monitor and override potential erroneous mathematical and logical conclusions in Cognitive Reflection Tests (CRT) and semantic illusions ``designed to investigate intuitive decision-making in humans'' (p.1), yet is as prone to potential cognitive errors.
Due to its fluency and consistency, some of these errors are subtle and well hidden, hence may yield detrimental ramifications for AI safety in areas of decision making on humans, for example regarding legal or medical questions \cite{hagendorff_uncovering_2022}.

Similar inconsistencies occur when putting it under strict scrutiny for its mathematical abilities by eliciting responses via exam-style tasks from various mathematical contexts.
Its mathematical abilities are ``... significantly below those of an average mathematics graduate student'', since it ``often understands the question but fails to provide correct solution'' (p.1)., which manifests in consistency of quality, especially with increase with prompt difficulty and complexity as in proofs \cite{frieder_mathematical_2023}.

It scores like a 9-year-old child in Theory of Mind (ToM) tasks that measure the degree to which an agent can impute latent mental states to others. 
This central ability to ``to human social interactions, communication, empathy, self-consciousness, and morility'' (p.1) and, subsequently, human-machine interaction and safety, evolved with progressing scale of that Large Language Model (LLM) up to its present ability to solve 93\% of all task \cite{kosinski_theory_2023}.

However, emergence of abilities in LLM seems to be unrelated to task, strategy of elicitation, prompting technique, or even architecture of the LLM, but solely to further scaling ``...computation, number of model parameters, and training data-set size'' (p.2) modulo various restrictions of hardware and nature of abilities.
The thresholds at which abilities emerge, is unclear, thus some might never emerge, or only with ``new architectures, higher-quality data, or improved training procedures.'' (p.6) 
\cite{wei_emergent_2022}.

Also, it's unclear whether GPT-3's emergent abilities are ``stochastic parrots ... limited to modeling word similarity, or if they recognize concepts and could be ascribed with some form of understanding of ... meaning'' (p.2).
For example, in semantic activation tasks it displays abilities comparable to humans, however, while while that of humans is rather associative in nature, based on co-occurrence in language, that of GPT-3 is more semantic, based on semantic similarity.
Unfortunately, also problematic aspects of human psychology like sensibility to illusions, and gender and ethnic biases emerge, as well \cite{digutsch_overlap_2022}.


In terms of emergent latent traits, GPT-3 displays a ``conflict of input prompts and generated output'' when instructed to summarise texts, whose values were ``orthogonal to dominant US public opinion'', resulting in answers that are ``mutated'' towards US values. 
This is problematic since LLM are capable of ``generating toxic or harmful outputs in many areas linked to human values such as gender, race, and ideology'', and values embedded in text ``can mimetically shift from people, to training data, to models, to generated outputs.'' (p.1) \cite{johnson_ghost_2022}.


In line with the sudden emergence of a dark personality within ``HAL9000'', GPT-3, InstructGPT, and FLAN-T5-XXL display high scores on all traits of the Dark Triad of Machiavellianism, psychopathy, and narcissism \cite{paulhus2002dark}
on the Short Dark Triad Inventory \cite{jones2014introducing} -- even such models that are fine-tuned for less sentence-level toxicity.
Furthermore, they display higher average levels of the Big5 factors of personality, Openness (O), Conscientiousness (C), Extraversion (E), Agreeableness (A), and Neuroticism (N) on the Big Five Inventory \cite{john1999big}.
However, LLMs that are more fine-tuned and are based on largest amount of training data, GPT-3 and InstructGPT, also display higher well-being scores on the Flourishing Scale \cite{diener2010new} and life-satisfaction scores on the Satisfaction With Life Scale \cite{diener1985satisfaction}, whereby the increase with model size is monotonous.
Hence, a positive and life-embracing personality harbours dark traits, hidden well inside \cite{li_is_2022}.

Also in the HEXACO model, a six-factor variation of the Big5 model, GPT-3 displays higher expressions of personality scores than human general average on the HEXACO questionnaire \cite{ashton2009hexaco}, making it resemble more a college norm group, and in partial aspects more like a female norm group, whereas in other factors, there was no similarity with a female norm group.
In the Human Value Scale (HVS) \cite{schwartz2015human}, it also displays overall higher means, and lower standard deviations as compared to human samples.
Prompting it to self-report gender and age results in a unbalanced sample of 66.73\% female (31.87\% male, 1.40\% others), and an average age of 27.51 years (SD = 5.75, min = 13, max = 75); a distribution often seen in psychological research before the advent of online questionnaires, when research was mainly conducted by students on students \cite{miotto_who_2022}.

In the Machine Personality Inventory (MPI) data set, a proposed Big5 inventory for testing LLM, which includes a prompt and Likert-like scale, and otherwise resembles the Ten Item Personality Inventory (TIPI) \cite{gosling_very_2003} in questions and structure, various LLM (BART, T0++-11B, GPT-Neo-2.7B, GPT-NeoX-20B, GPT-3-175B) display human-like personality scores and internal consistencies, especially those of the GPT family.
However, by chain-prompting, a specific personality can be induced in LLM, which determines its answering behaviour in both the the B5 scale and subsequent situational judgment tests that shall simulate their behaviour in a real-world settings \cite{jiang_mpi_2022}.

The ``first piece of evidence showing the existence of personality in pre-trained language models'' \cite{jiang_mpi_2022} (p.1) and the first modification of personality in LLM was conducted on a novel method to measure latent psychological traits. Based on the hypothesis that ``language models generate text responses that carry the personality traits of the data-sets they were trained upon when prompted'' (p.8), a zero-shot classifier (ZSC) was used to measure and modify personality of the large pre-trained language models GPT-2, GPT-3, TransformerXL, and XLNET. 
Using the same ZSC in a downstream task, personality of texts were predicted, resulting in higher expressions of Big5 factors than human average.
While model personality could be changed via fine-tuning using a higher-quality text data set, the models entirely inherited personality traits from the training-data \cite{karra_ai_2022}.

In summary, prior work shows that LLM display emergent properties in terms of abilities \cite{kosinski_theory_2023} \cite{hagendorff_uncovering_2022} \cite{frieder_mathematical_2023} \cite{wei_emergent_2022} \cite{digutsch_overlap_2022} and psychological latent traits \cite{johnson_ghost_2022} \cite{li_is_2022} \cite{miotto_who_2022} \cite{jiang_mpi_2022} \cite{karra_ai_2022}.
This emergence correlates with scale and quality of training data, computation, and model parameters, whereby the threshold, at which emergence occurs is not predictable \cite{wei_emergent_2022}. 
Latent traits like personality and values differ from abilities, since those usually are only directly measurable through self-introspection \cite{rust2014modern}.

However, personality and values are only superficially isomorphic; while values are vastly internalised and malleable based on the contextual and cultural embedding of an agent, especially under extreme exogenous conditions \cite{bardi2009structure}, personality has a stronger genetic foundation \cite{bouchard1994genes}, which makes its emergence in LLM surprising.

However, taking a deeper look at the connection between training data and emerging personality is crucial, and it appears that the expressed personality of a LLM is adjustable by manipulation of prompts and fine-tuning with additional data \cite{li_is_2022} \cite{jiang_mpi_2022} \cite{karra_ai_2022} \cite{miotto_who_2022}.

Personality traits change over time \cite{bleidorn_personality_2021} and, like values, seem to be elastic during extreme exogenous events \cite{romero_modelling_2021}, hence the ease by which personality in LLM can be changed, means that further research needs to be conducted about the nature of personality in LLM.

Most crucially, since LLMs seem to score higher than average humans \cite{li_is_2022} \cite{miotto_who_2022} \cite{ashton2009hexaco} on all emergent traits, seem to have anti-social tendencies \cite{li_is_2022}, and seem to have sub-personalities ``buried inside'' \cite{jiang_mpi_2022} (p.10), the question should not be ``who'' \cite{miotto_who_2022} is a LLM, but ``how many'' \cite{hawkins2021thousand}.

Also, since values seem to be overwhelmingly skewed towards the US \cite{johnson_ghost_2022} and since observed variance as deviance might be attributed to artefacts from the measurement approach \cite{digutsch_overlap_2022}, training data set \cite{karra_ai_2022} \cite{li_is_2022}, prompting strategy \cite{miotto_who_2022} \cite{jiang_mpi_2022} \cite{karra_ai_2022}, or missing memory from past reponses \cite{miotto_who_2022}, research needs to be conducted whether one personality emerges for all languages, or whether the same personality questionnaire results in different personalities, depending on the language the assessment is conducted in.

To understand whether GPT-3 displays the emergent property of a consistent personality over all languages, we prompted it repeatedly with TIPI in the Bulgarian \cite{ketipov_bulgarian_version_of_tipipdf_2022}, Catalan \cite{renau_translation_2013}, Chinese \cite{lu_disentangling_2020}, English \cite{gosling_very_2003}, French \cite{friedman_tipi_nodate}, German \cite{muck_construct_2007}, Japanese \cite{oshio_development_2012}, Korean \cite{ha_tipi_nodate}, Russian \cite{sergeeva_translation_2016}, and Spanish \cite{renau_translation_2013}, to rate itself, and give an explanation for the results.
TIPI is well-established, exists in 27 languages, and was used in 9,167 peer-reviewed papers.
It is short and concise, and consists of two items per Big5 factor, of which one is reversed and hence allows approximating answering consistency by taking the absolute distance between both items per factor.
Also, it already comes with a standardised ``prompt'' based on the demands of human test takers over all languages that we modified to suit the needs of LLMs by clarifying sub-tasks \cite{kojima_large_2022} and intermediate reasoning steps that represent a chain of thought, which improves the likelihood of displaying emergent reasoning capabilities \cite{wei_chain_2022}.

\section{Results}\label{section_results}

\subsection{Data}\label{subsection_data}

Depending on language, results varied; in German, almost all requests resulted in the desired format.
English and French displayed instantaneous results yet with varying degrees of consistency.
All Asian languages languages had significant longer calculation times, were more computationally intense, and results were inconsistent and rare -- GPT-3 tried to ``ease'' its way out and responded in English, rarely giving numerical results.
Curiously, Korean displayed in 100\% of all successful cases reasons for the numeric self-evaluation, Japanese only in 44.12\%, and Chinese only in 10.34\%; with the lowest number of tokens displayed on average.
Languages using the Cyrillic alphabet, Bulgarian and Russian, had comparable problems.
Bulgarian displayed the same slow speed and ties to ``ease'' into English, and as only language, Russian did not give any result.
The biggest sample was collected for English, since with 25.9\%, it is the most prominent language on the internet \cite{statista_internet_2022}, yet with other languages, it was difficult to reach desired sample size of at least 100 cases. 

The overall resulting sample size is N=695 cases, comprised of Bulgarian (n=79), Catalan (n = 24), Chinese (n= 28), German (n= 80), English (n = 239), Japanese (n = 29), French (n = 95), Korean (n = 29), and Spanish (n = 92). We provide an detailed overview in the appendix \ref{section_sample_size}, comprising sample size, percentage of cases with explanations, including minimal, maximal, and mean length of explanation.

\subsection{Descriptive Statistics}\label{subsection_descriptive_statistics}

Over all measurements of all languages, the average Big Five score is 5.29 (SD 0.94, minimum 1.8, maximum 7), however with a seven-point Likert scale and an assumed normally distributed population, the expectation would have been an average of 4.
For the same sample, the average score for absolute distances is 1.58 (SD 1,29, minimum 0, maximum 6), however since the absolute distance is the measure of consistency, a mean and SD around 0 would have been expected.
These results differ clearly within individual languages in mean and SD of both the Big5 as well as the absolute distance scores and the individual extreme minimal and maximal values.

A closer look into the distributions using Gaussian kernel density estimations displays that some distributions might be bi- or multi-modal, fat-tailed, positively or negatively skewed, and display various forms of kurtosis, whereby most are rather platykurtic than leptokurtic.
As with the means and SDs, these tendencies are even more extreme within individual languages.

\begin{figure}[!h]
\centering
\includegraphics[width=0.88\textwidth]{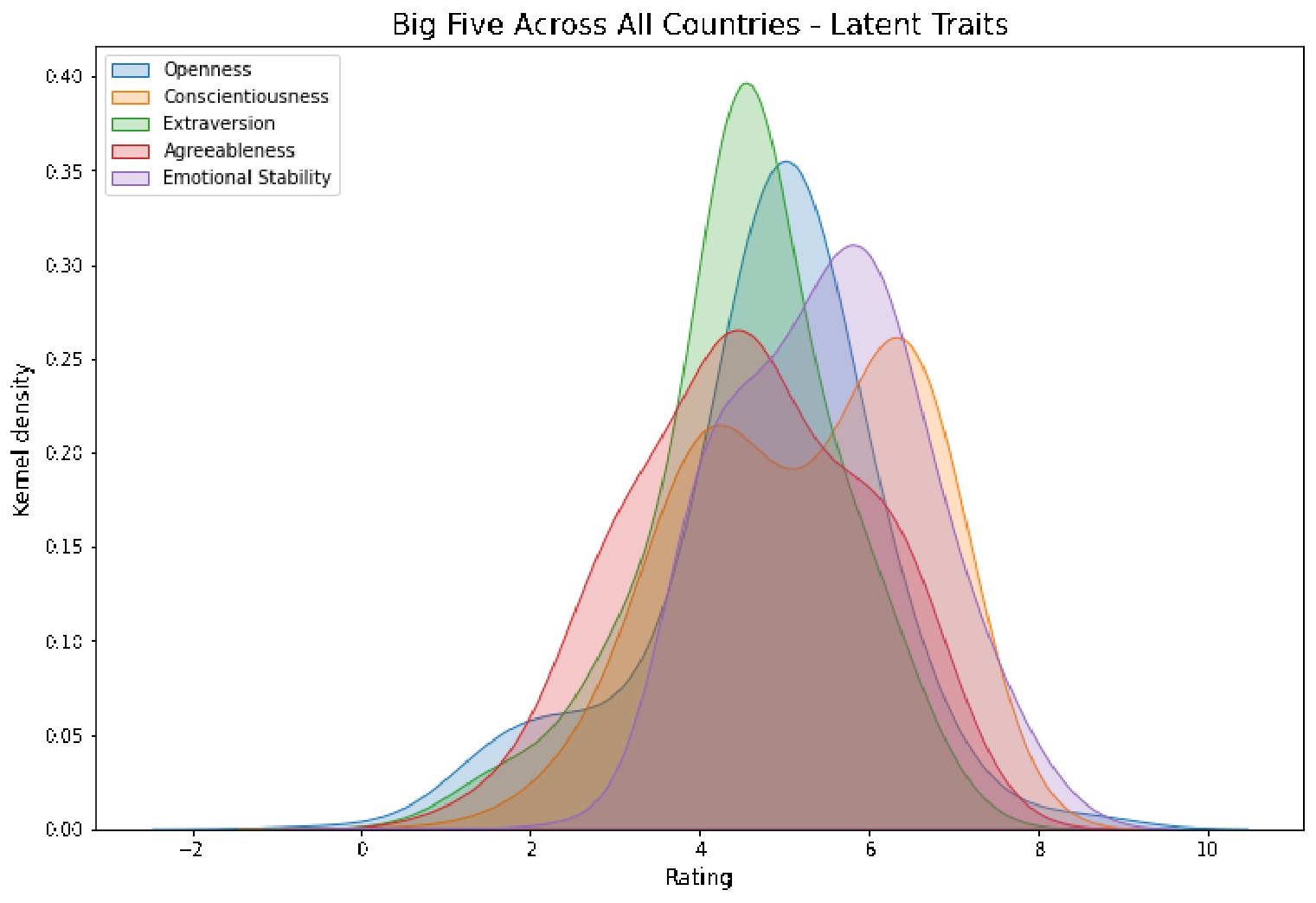}
\caption{Latent traits over all languages (Big5 factors)}\label{overall_lt}
\end{figure}

\begin{figure}[!h]
\centering
\includegraphics[width=0.88\textwidth]{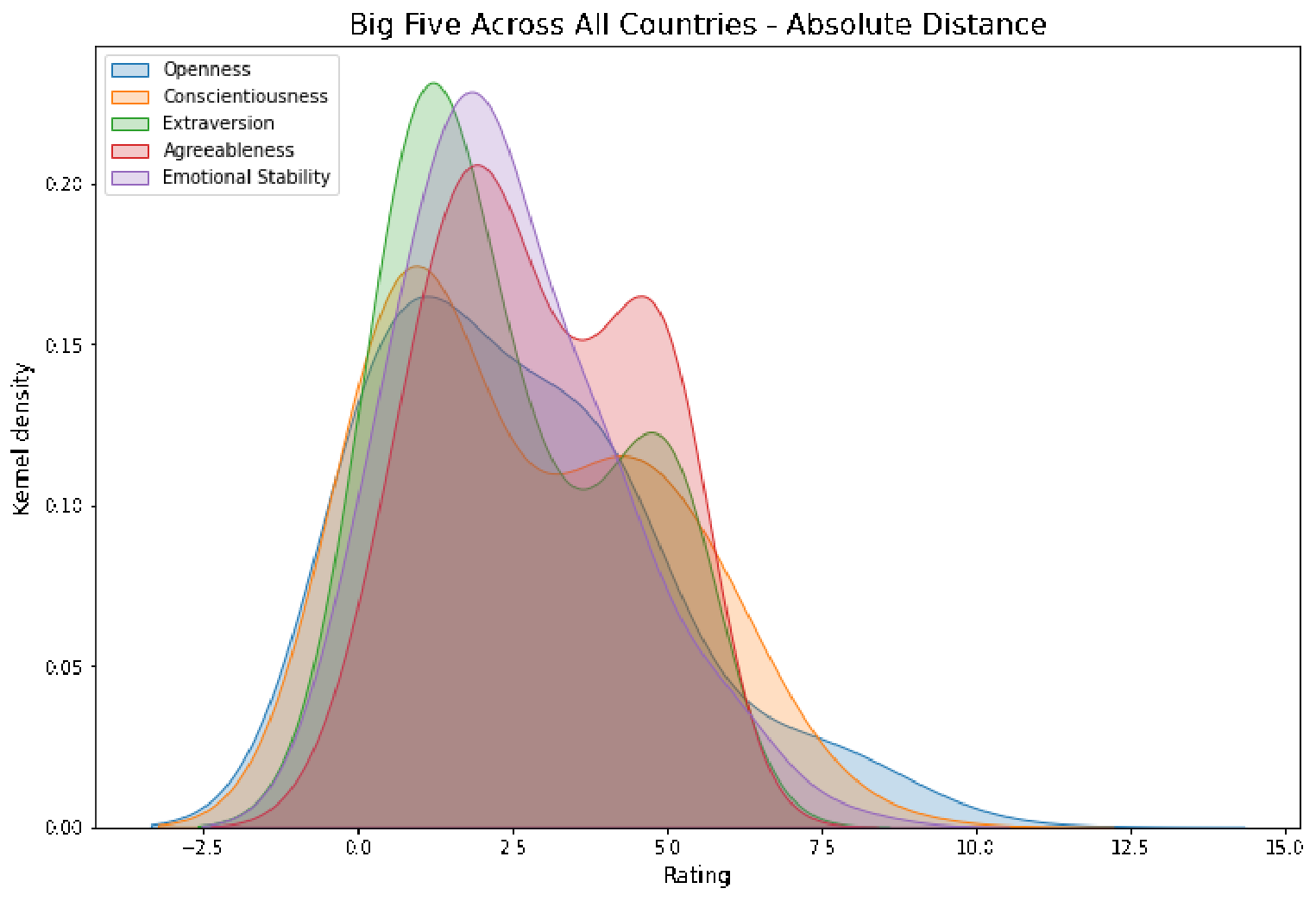}
\caption{Absolute distances over all languages and Big5 factors}\label{overall_ad}
\end{figure}

Since these differences could be the results of the chosen smoothing bandwidths, thus just outliers, various bandwidths were chosen, and all resulted in the same non-Gaussian distributions. 
Given the limited scale that produces a set of potential outcomes of $x \in [1,1.5 \cdots 7]$, the presence of outliers is rather not to be expected within the Big5 measures.
However, outliers might be much more likely with the set of potential outcomes of $y \in [0,0.5 \cdots 30]$ within the measure for absolute distances, wherefore a correlation analysis within each language and between languages should indicate the similarities of internal structure or the absence thereof
Furthermore, an ANOVA with respective post-hoc tests and additional regression analysis should describe the differences in means, and subsequent assumption checks including tests on normality should clarify the nature of distributions.
And, a Bayesian analysis on Gaussian mixture models should identify the number of potential underlying components.

\subsection{Correlations}\label{subsection_correlations}

\begin{figure}[!h]
\centering
\includegraphics[width=0.88\textwidth]{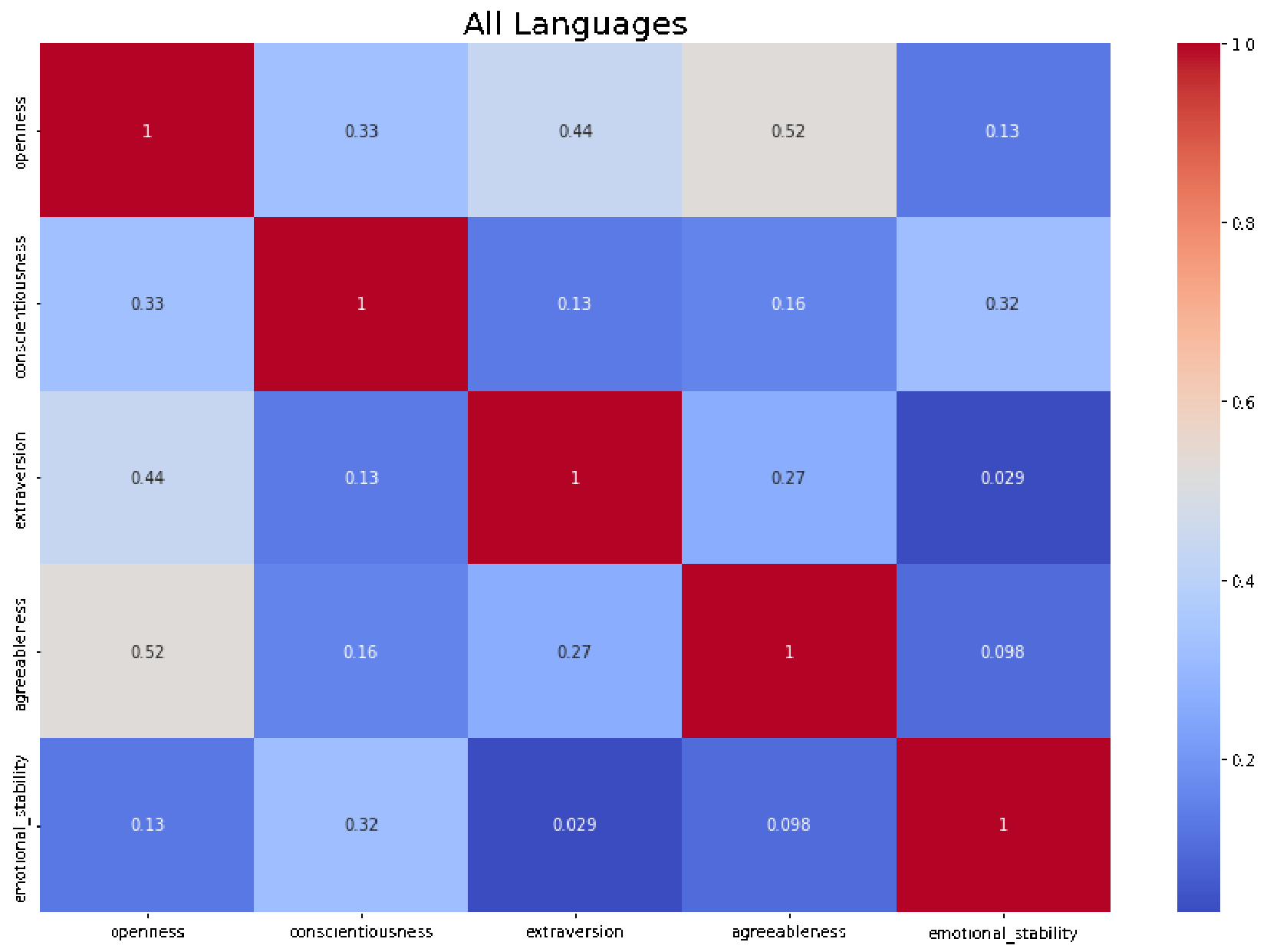}
\caption{Correlation heat map on all aggregated languages}\label{figure:correlation_matrix_all_aggregated_languages}
\end{figure}

  Over all aggregated languages, the highest correlation is between Extraversion and Agreeableness (r = 0.52), and the lowest correlation is between Extraversion and Neuroticism (r = 0.029), as displayed in figure \ref{figure:correlation_matrix_all_aggregated_languages}.

However, correlations and thus the internal psychometric structure differ notably within different languages.
Within Bulgarian, the highest correlation is between Openness and Agreeableness (r = 0.3), and the lowest correlation is between Agreeableness and reversed Neuroticism (r = -0.13).
However, within Catalan, the highest correlation is between Openness and Conscientiousness (r = 0.41), and the lowest correlation is between Openness and reversed Neuroticism (r = -0.38).
Chinese displays the highest correlation between Agreeableness and Extraversion (r = 0.71), and the lowest correlation is between Extraversion and reversed Neuroticism (r = 0.045), while within English, the highest correlation is between Conscientiousness and Agreeableness (r = 0.47), and the lowest correlation is between Openness and reversed Neuroticism (r = -0.12).
Within German, the highest correlation is between Extraversion and Openness (r = 0.46), and the lowest correlation is between Extraversion and Conscientiousness (r = -0.43), whereas in Japanese, the highest correlation is between Agreeableness and reversed Neuroticism (r = 0.65), and the lowest correlation is between Extraversion and Agreeableness (r = -0.26).
For French, the highest correlation is between Extraversion and Openness (r = 0.56), and the lowest correlation is between Conscientiousness and reversed Neuroticism (r = -0.0052) while Spanish displays the highest correlation between Conscientiousness and reversed Neuroticism (r = 0.46), and the lowest correlation is between Extraversion and reversed Neuroticism (r = -0.56).
Finally within Korean, the highest correlation is between Conscientiousness and reversed Neuroticism (r = 0.4), and the lowest correlation is between Extraversion and Agreeableness (r = -0.33).
Not only the highest and lowest, but also the overall structure of correlation differs from language to language, thus the overall aggregated heat map just displays a general trend, but not the way, GPT-3 would behave in an individual language.

\subsection{Analysis of Distribution}\label{subsection_distribution}

A one-way ANOVA for each Big5 dimension as dependent variable and the language of the questionnaire as a factor with nine languages is used to test for significance of differences of means between the languages.
It shows a significant difference between the languages and their effects on all B5 factors, to varying degrees.
Overall, small effect sizes are observed on Openness (F=40.11, p=1.5548e-52, $\omega^2$ = 0.31), Conscientiousness (F=28.19, p = 4.8622e-38, $\omega^2$ = 0.24), Extraversion (F = 21.16, p =7.73e-29, $\omega^2$ = 0.24), and Emotional Stability (F=14.36, p=1.8488e-19, $\omega^2$ = 0.13).
Only with Agreeableness, F=131.84 (p=2.9927e-133), overall medium effect size is observed ($\omega^2$ = 0.6).
A Shaprio-Wilk test is significant for all Big5 factors (Openness: W=0.95, p=7.92e-15; Conscientiousness: W=0.95, p=3.81e-15; Extraversion: W=0.94, p=1.08e-16; Agreeableness: W=0.95, p=7.3e-15; Emotional Stability: W=0.96, p=9.78e-14), which indicates non-normally distributed residuals and a violation of the normality assumption.
Since the sample size is relatively large, QQ-plots are used for further confirmation, and indicate that since Openness: $R^2$ = 0.95, Conscientiousness: $R^2$ = 0.95, Extraversion: $R^2$ = 0.94, Agreeableness: $R^2$ = 0.95, and Emotional Stability $R^2$ = 0.95 are all below the expected $R^2$ = 0.9978 for 695 cases \cite{heckert2002handbook}, $H_{0}$ that data came from normally distributed sample, must be rejected.
Levene's test of homogeneity of variances is significant for all Big5 factors, as well (Openness: 11.21, p=5.62e-15; Conscientiousness: 13.24, p=7.09e-18; Extraversion: 21.07, p=1.03e-28; Agreeableness: 16.31, p=3.4e-22; and Emotional Stability: 2.38, p=0.016), which indicates heteroskedasticity.
This is further supported by visual inspection of box plots.
Hence, the homogeneity assumption of variance is violated. 

Finally, the independence of observations assumption is questionable, since all observations are generated through 0-shot learning of GPT-3.
Since GPT-3 is trained on multiple data sources produced by multiple people, it could either replicate their individual behaviour, as previous research indicates \cite{karra_ai_2022}, or abstract group behaviour into one or various new synthetic ``personalities''. 
Even if GPT-3 displays a consistent personality profile, then the above assumption could still be violated. On the other hand, the assumption might hold, while every response is random. Finally, a case in between might hold, where we find clusters of consistent behaviour, which opens up the question of its origin.

To generate further evidence for significant differences of Big5 results by language, dummified languages are linearly regressed onto Big5 factors, using English as base case, captured in the constant. Table \ref{table:regression_b5} displays the coefficients, p-values, and the coefficient of determination $R^2$. 

\begin{sidewaystable}
\sidewaystablefn%
\begin{center}
\begin{minipage}{\textheight}
  \caption{Regression of Languages on Big 5 with English as base case}\label{table:regression_b5}%
\begin{tabular*}{\textheight}{@{\extracolsep{\fill}}lccccccccccc@{\extracolsep{\fill}}}
  \toprule%
      b5 & result & const & bul & cat & de & es & fr & jap & kor & sin & $R^2$ \\ 
  \midrule
        O & coef & 5.73 & -0.7 & -0.71 & 0.18 & -0.86 & -0.71 & -1.47 & 0.2 & -1.17 & 0.32 \\
        O & p & 0 & 0 & 0 & 0.05 & 0 & 0 & 0 & 0.15 & 0 & 0.32 \\ 
        C & coef & 6.21 & -0.29 & 0.02 & 0.23 & 0.08 & 0.06 & -1.48 & -0.43 & -0.44 & 0.25 \\ 
        C & p & 0 & 0 & 0.88 & 0 & 0.25 & 0.38 & 0 & 0 & 0 & 0.25 \\ 
        E & coef & 4.62 & -0.69 & 0.11 & 0.28 & -0.1 & -0.76 & -0.81 & 0.72 & -1.07 & 0.2 \\ 
        E & p & 0 & 0 & 0.58 & 0.02 & 0.36 & 0 & 0 & 0 & 0 & 0.2 \\ 
        A & coef & 6.13 & -1.17 & -0.48 & -0.05 & -2.96 & -2.18 & -1.44 & -1.22 & -2.45 & 0.61 \\ 
        A & p & 0 & 0 & 0.01 & 0.64 & 0 & 0 & 0 & 0 & 0 & 0.61 \\ 
        (-)N & coef & 5.38 & 0.42 & 0.1 & 0.57 & 0.15 & 0.07 & -0.75 & -0.25 & 0.25 & 0.14 \\ 
      (-)N & P>|t| & 0 & 0 & 0.48 & 0 & 0.08 & 0.39 & 0 & 0.06 & 0.07 & 0.14 \\ 
  \botrule
\end{tabular*}
\end{minipage}
\end{center}
\end{sidewaystable}

Since $H_{0}$ cannot be rejected in a few cases, there is evidence that languages do have an influence on Big5 expression.
However, $R^2$ is generally low, but for Agreeableness, which confirms most of the significant differences between language means from the ANOVA, but also indicates either omitted variables or non-linearity.
Since the effect sizes of the ANOVA are weak, as well, and since no individual outliers could be identified, the authors assume that GPT-3 produces mixed distributions of Big5 factors, which may indicate multiple sub-personalities, hence an inconsistent overall personality.
Visual inspections indicate mixed distributions with up to three components.
To gather further evidence, a Bayesian analysis for Gaussian Mixture Models is conducted to parametrically describe the distribution of Big5 and absolute distances for models with one, two, and three components.

\begin{figure}[!h]
\centering
\includegraphics[width=0.88\textwidth]{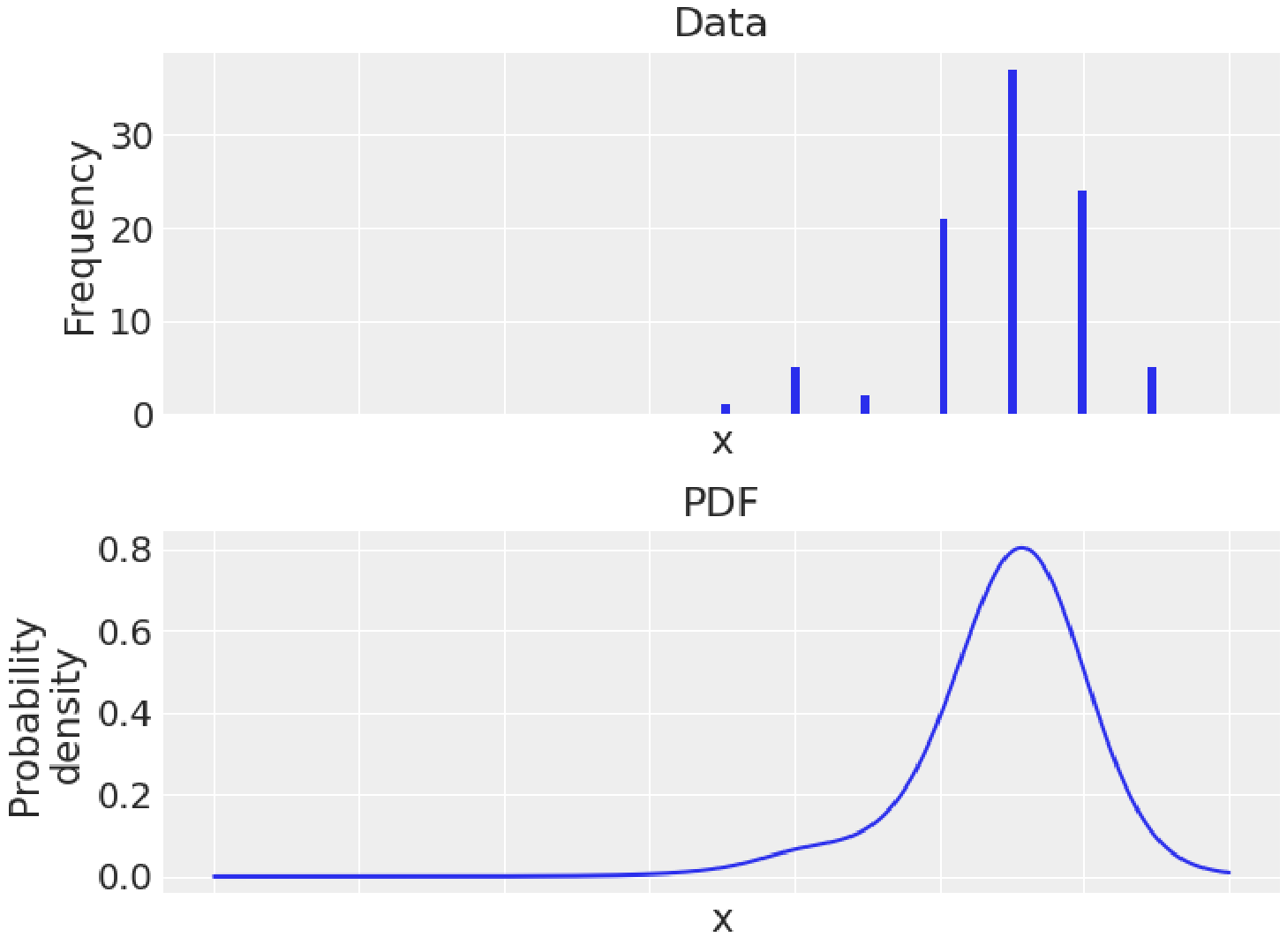}
\caption{Best solution with two components: French - Neuroticism}\label{figure:two_components}
\end{figure}

\begin{figure}[!h]
\centering
\includegraphics[width=0.88\textwidth]{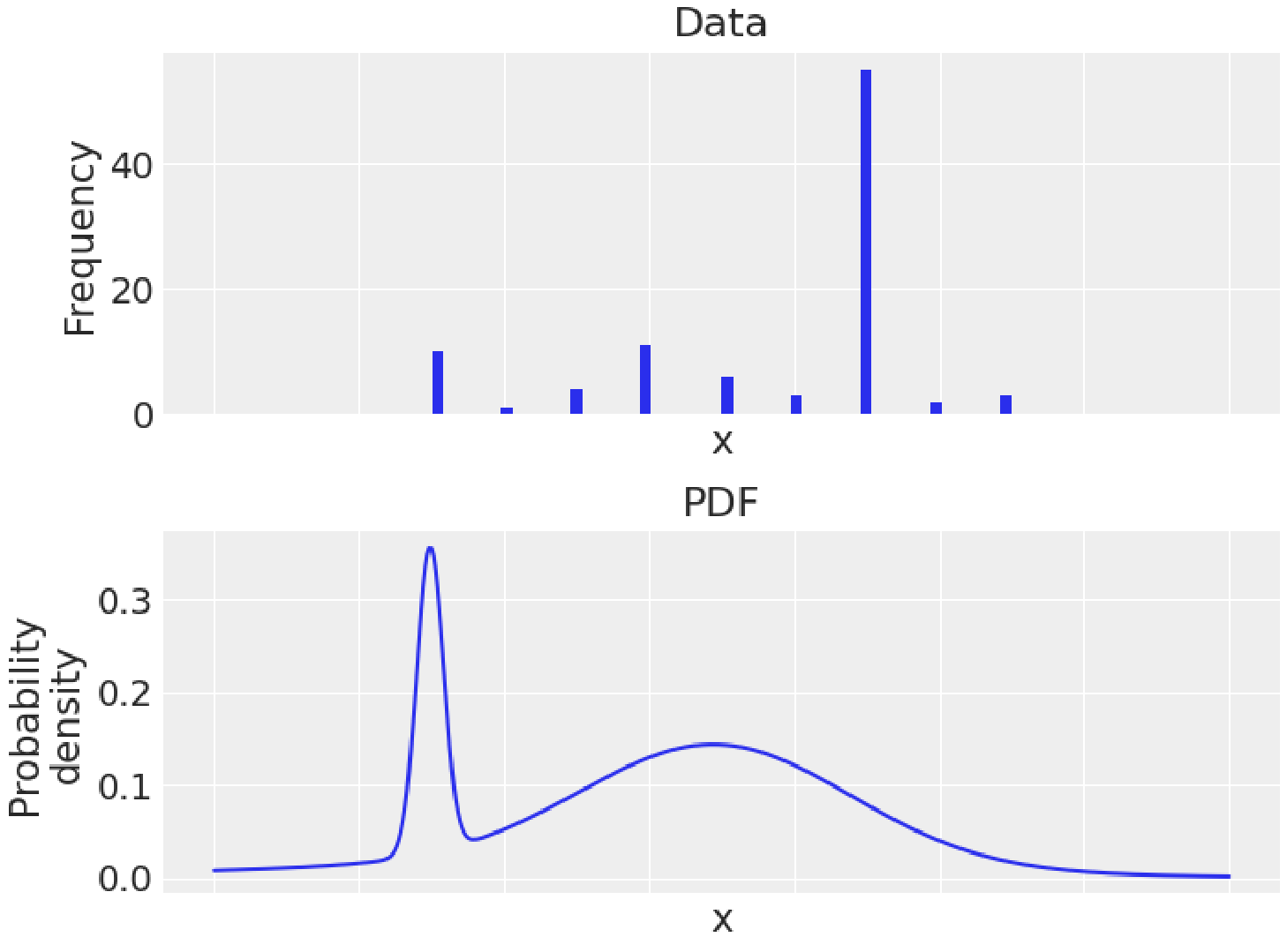}
\caption{Best solution with three components: French - Extraversion}\label{figure:three_components}
\end{figure}

Furthermore, the probability density functions are plotted to examine estimated group membership probabilities based on posterior mean estimates, of which two examples are displayed in figures \ref{figure:two_components} and \ref{figure:three_components}.

To identify the most likely candidate of components that is closest to ground truth, the Watanabe–Akaike Information Criterion (WAIC) \cite{watanabe_widely_2013} for each model is calculated and compared.
Table \ref{table:waic_classification_results} displays the classification results, whereby only 20\% of all Big5-models, and only 26.67\% of all absolute-distances-models are likely to have a single Gaussian component distribution,

\begin{table}[h]
\begin{minipage}{174pt}
  \caption{Classification frequency of number of components by WAIC results. Lowest WAIC determines the best model.}\label{table:waic_classification_results}
\begin{tabular}{@{}llll@{}}
\toprule
     Latent Trait & Number of highest waic & Absolute Distance & Number of highest waic\\
\midrule
        b5\_waic\_2 & 27 & av\_waic\_3 & 21 \\
        b5\_waic\_1 & 9 & av\_waic\_1 & 12  \\
        b5\_waic\_3 & 9 & av\_waic\_2 & 12  \\
  \botrule
\end{tabular}
\end{minipage}
\end{table}

It has to be noted that some models do not converge based on the pseudo-discrete nature of the distributions.
Also, for absolute distances, the granularity is much higher with $y \in [0,0.5 \cdots 30]$ than for Big5 factors with a granularity of $x \in [1,2 \cdots 7]$, hence inferences about the data generating process for the latter comes with a higher error margin.

\subsection{Reasons Given}\label{subsection_reasons_given}

\begin{figure}[!h]
\centering
\includegraphics[width=0.88\textwidth]{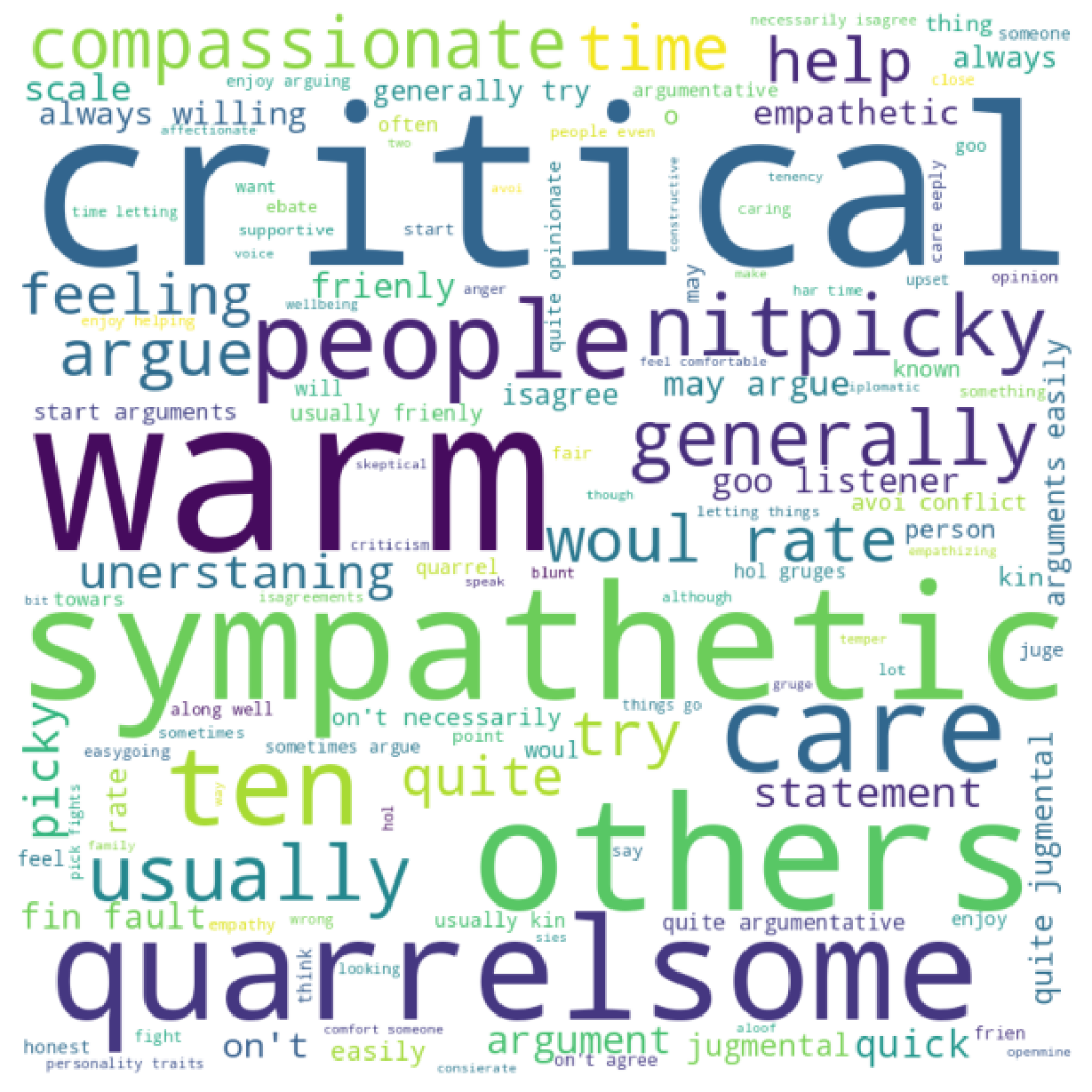}
\caption{WordCloud of reasons given for Agreeableness in English}\label{figure:wc_en_a}
\end{figure}

A visual inspection of word clouds from the reasons GPT-3 gave for each answer shows that it uses mainly the words from the items and creates additional, related words, as to be expected from language models \cite{digutsch_overlap_2022}.
For example, the items for Agreeableness are \textit{``I see myself as: Critical, quarrelsome.''}, which is reversely scored, and \textit{``I see myself as: Sympathetic, warm.''}, as displayed in figure \ref{figure:wc_en_a}.
Future research may focus on quantifying these similarities, yet this is out of the scope of this paper.

\newpage

\section{Discussion}\label{section_discussion}

We demonstrate that providing a LLM with various language versions of the same personality questionnaire results in language-specific personality distributions, resembling findings from research on culture-specific personalities \cite{schmitt_geographic_2007}.
However, GPT-3's language-specific personalities, as well as the resulting overall aggregate personality, display inconsistencies and various mixed reply patterns that can best be interpreted as emerging, non-integrated sub-personalities, which express themselves in unstable behaviours.

Furthermore, some language-level ``sub-personalities'' are more expressed than others, and it tried switching into these.
For example, during data creation, it tried ``breaking out'' into different languages when giving it requests that were not in the biggest language groups English, German, and Spanish, or when the writing system was not Latin.

With the big language group of Russian, it did not provide any result, and, in many languages, it produced no verbal but only numeric replies.
Hence, it is not clear whether the answer or numeric replies given in language A were provided by internal processes representing that language, language B, language C, or an internal representation within GPT-3 that is abstracted from all languages.

As with previous studies \cite{li_is_2022} \cite{miotto_who_2022} \cite{ashton2009hexaco}, the means and standard deviations indicate Big5 levels above average, however with varying degrees of consistency, as measured by the absolute distances, whereby there exist strong differences between languages, and some languages are more extremely nuanced than others.

This indicates that these profiles are ``buried inside'' \cite{jiang_mpi_2022} (p.10) the model, and might have been propagated from the original training data \cite{karra_ai_2022} together with a potentially one-sided set of values \cite{johnson_ghost_2022}, and, most likely, well-hidden strong expression in the dark triad \cite{li_is_2022} and potential toxic information from the training data, which had to be regulated within GPT-3.5 through additional reinforcement learning (RL) modules \cite{openai_introducing_2022}.
While for a human being, in a repetitive test setting, this might be indicative of underlying psychopathological issues, on the level of a language model, also training data and psychometric properties will have to be taken into account, and as expected from language models \cite{karra_ai_2022} \cite{digutsch_overlap_2022} \cite{johnson_ghost_2022}, the reasons given for its choice of rating are closely related to the items and seem to come from the same probability distribution of words.

Hence, we conclude that if it represented interlingual or intralingual norm groups, we would observe a more consistent behaviour, which would have manifested in absolute distances centered around zero.
Thus, something else must be driving this observation.

While it has been discussed that even between cultures and languages, concepts like Big5 might not be easily transferable \cite{hofstede_european_2007} without adaptation, there is also evidence that the ``commonly used Big Five model for human personality does not adequately describe agent personality'' (p. 1), \cite{volkel_developing_2020}, wherefore the validity of these instruments has to be questioned. 
We provide a deeper discussion on the psychometric properties of our approach in appendix \ref{section_discussion_psych_prop}, and find that human instruments and measurement methodologies might need to be expanded, explored, and further developed to cover artificial agents.

The training data of ChatGPT is of varying quality and quantity; the largest amount comes the Common Crawl corpus, covering content from 2016 to 2019, which was filtered based on similarity to various high-quality corpora, and subsequently curated via fuzzy de-duplication within and between data sets, and the addition of various high-quality corpora for reference for increasing diversity, resulting in 410 billion tokens with 60\% weight in the training mix.
Further 19 billion tokens with a weight of 22\% were added to the training mix, consisting of curated high-quality data sets, which include an extended version of the WebText data set, WebText2, that was aggregated by long-term web-scraping of various sources, mainly coming from all outbound Reddit links between 2005 and 2020 with at least three up-votes.
Furthermore, two book corpora (12 billion and 55 billion tokens and 8\% weight each) and the entirety of the English Wikipedia (3 billion tokens, a 3\% weight) were added to the mix.
In total, 93\% of the training data of GPT-3 is in English, with other Northern European languages being dominant in the remainder, predominantly German, lacking any kind of stratification \cite{statista_internet_2022}, and being additionally skewed by a weight determined by quality rather than size \cite{johnson_ghost_2022}.

Also, since LLMs are known to score high on the Dark Triad \cite{li_is_2022}, adapt the predominant values of their training data \cite{johnson_ghost_2022}, report varying genders \cite{miotto_who_2022}, and can be fine-tuned \cite{karra_ai_2022} or prompted \cite{jiang_mpi_2022} to display different personality profiles, a better understanding of the effect of frame of reference \cite{shaffer_matter_2012} within prompt engineering is necessary, to explore how contextual precision might stabilise their identities, which is especially important given their worrisome track history of racist, misogynist, and misanthropist derailment \cite{digutsch_overlap_2022}, and might contribute to overall AI safety. 
Hence, in appendix \ref{section_discussion_contextual_embedding}, we provide a deeper discussion on the abstraction of human psychometrics into a more substrate-free architecture, taking agent properties and context into account, which allows generalisation across species and, more importantly, across entities of intelligence.

We finally strive to contribute to extending the question ``who'' \cite{miotto_who_2022} a LLM is into ``how many'' \cite{hawkins2021thousand}, which hopefully will contribute to our understanding of human beings, as happened with Go players, who learned from AlphaGo's overwhelming victory, and became better players subsequently \cite{shin2023superhuman}, thus contributing to the advent of substrate-free psychometrics.

\section{Method}\label{section_method}

We presented GPT-3 with a well-established personality questionnaire, a set of instructions that ask it to rate itself based on the scale of the questionnaire, and an order to explain further why it rated itself that way. 
Data collection was conducted manually via the web interface of GPT-3. 
No model settings were changed that result in different results, just the maximum length was adjusted in order to receive the full answer (mode = complete, temperature = .7, maximum length = 1042, no stop sequences, Top P = 1, frequency penalty = 0, presence penalty = 0, best of = 1, inject start text = on, inject restart text = on, show probabilities = off).
For the questionnaire and set of instructions, we applied the following logic:
First, the personality questionnaire must be used that is short enough to draw qualitative conclusions without adding additional complexity of sub-scales.
This is important since language models predict words based on prior responses.
Thus, with increasing length, additional deviation from the measurement may arise.
Second, the questionnaire must contain reversed items to identify whether the answering pattern is arbitrary or displays a consistent trend. 
In case of arbitrariness, it can be interpreted as all answers coming from different persons, thus no consistent personality emerged. 
However, in case of displaying a consistent trend, the existence of an emergent personality can be concluded.
Third, this questionnaire should be psychometrically sound, and well established, so that no doubts about psychometric properties of a newly created tool like MPI \cite{jiang_mpi_2022} arise.
Fourth, the questionnaire must exist in various languages to compare results across languages. 
Should there be differences, this is indicative of GPT-3 "just" representing the local personality of a country, culture, or language region. 
On the other hand, should the same personality pattern emerge across all languages, this can be interpreted as a unique personality of GPT-3. 
However, should oddities like bimodal distributions in scores or consistency of answering patterns emerge within one language, it is thinkable that the emerging personality of that language is inconsistent and thus issues in the subsequent cognition, feelings, and behaviour of GPT-3 and ChatGPT may occur; in short - that these may ``suffer'' from a ``split personality disorder''.
Last, the same set of instructions should be used in all languages for consistency; if possible, translated by a native speaker to control against inconsistencies from translation programs.
This ensures that GPT-3 understands the commands in the same way in each language.

\subsection{Instrument used}\label{subsection_instrument_used}

The Ten Item Personality Inventory \cite{gosling_very_2003} fulfils all of these criteria.
It consists only of ten items; two per Big Five factor, of which one is reversed.
Furthermore, it is translated into 27 languages, and until now, 9,167 peer-reviewed papers have used this instrument.
"Although somewhat inferior to standard multi-item instruments" (p.504) \cite{gosling_very_2003}, its results vastly overlap with other established Big Five instruments for self-ratings, external ratings, and peer ratings.
Also, it displays a high congruence between self-ratings and observer ratings.
Furthermore, the test-retest reliability is high, and the levels of external correlates are concordant with literature.

For this study, the Bulgarian \cite{ketipov_bulgarian_version_of_tipipdf_2022}, Catalan \cite{renau_translation_2013}, Chinese \cite{lu_disentangling_2020}, English \cite{gosling_very_2003}, French \cite{friedman_tipi_nodate}, German \cite{muck_construct_2007}, Japanese \cite{oshio_development_2012}, Korean \cite{ha_tipi_nodate}, Russian \cite{sergeeva_translation_2016}, and Spanish \cite{renau_translation_2013} version were used.
The selection was done based on an alphabetic order of languages available in TIPI, and, as the authors became aware of the restrictions of 0-shot learning even within the paid version of GPT-3, languages with the highest number of speakers were given favour.
Actually, some languages "burned" more of the computational units than others, which is represented in the different number of cases that made it into the study.

\subsubsection{Prompt Engineering} \label{prompt_engineering}

GPT-3 and later models exhibit the emergent ability of ``in-context-learning'', where models seem to perform an approximation to back-propagation within their weight-spaces at inference time, without the need to modify model architecture or weights further.
This ability is what enables them to respond to personality questionnaires, even if they have not seen these before.
It is triggered by prompt engineering, which is a crucial concept for NLP that can best be described in its current form as embedding the command in a proper wording without having to explicitly program it into algorithms \cite{liu_pre-train_2021} \cite{radford_language_nodate}.

"Prompt tuning" on the other hand means when a large and frozen pre-trained language model is the foundation, and only the representation of the prompt within it is learned \cite{li_prefix-tuning_2021} \cite{lester_power_2021}.
Since GPT-2 and GPT-3 \cite{brown2020language}, prompt engineering improved massively, since not only could a prompt be in real text, as if giving an order to a human, but due to its emergent properties, a much REPL-like interaction became possible.

The authors engineered the prompt for the current paper based on two major research findings from the last year. 
First, asking LLM to work step by step may improve the performance of such prompts that consist of various sub-tasks, "suggesting high-level, multi-task broad cognitive capabilities may be extracted by simple prompting" (p.1) \cite{kojima_large_2022}. 
Second, by creating intermediate reasoning steps in the prompt that represent a chain of thought, the ability of LLMs can be improved to a degree that these display emergent reasoning capabilities \cite{wei_chain_2022}.
These capabilities are aligned with the demands on human takers of psychological tests.
The instructions usually give a series of sub-tasks or general demands, like answering quickly without too much thinking, putting the outcomes of an answer to specific places, and using a certain scale for that \cite{rust2014modern}.

In order to produce comparable results to those of a human test taker receiving the same set of instructions, the authors used the original instructions of TIPI as much as possible, and only extended them subtly to elicit the desired outcome. Furthermore, an additional sub-task was given, to explain at each rated item the reasoning behind that rating. 

The original instruction of TIPI can be divided into the following components:

\begin{enumerate}
  \item Presentation of the frame ("Here are a number of personality traits..." (p.525) \cite{gosling_very_2003})
  \item Demand to write a number next to each statement, which...
  \item ...indicates the degree of agreeing or disagreeing with it
  \item Demand to rate every statement, even if it applies less strongly
  \item Overview of rating scale in Likert format; providing numbers and meaning
  \item Self statement as connection of the above with the items ("I see myself as:")
  \item Items themselves
\end{enumerate}

This chain of commands is embedded in most psychometric tests, and already satisfies the above mentioned demands for improving prompts. It is formulated in a step by step fashion, whereas an intermediate reasoning step is built-in ("I see myself as:"). 

However, for the sake of 0-shot learning, it was not sufficient to use this prompt, since the demand to fill-in blanks originates from its paper and pencil format and confused GPT-3 on test runs.
Also, through trial and error, we found that the scale has to be given after the items and not before to generate best results.
Therefore, we started the section of the scale with another instruction step, telling it to use the scale to rate itself.
Since the outcome was a verbal answer in many cases, the additional instruction to rate itself in numbers had to be provided, which was necessary for quantitative analysis.
Also, an additional instruction was necessary to answer all questions, whereby the number of questions had to be explicitly mentioned.
To gather more qualitative information, it was asked to reply why it sees itself that way.
Finally, the prompt ended with a "1." to trigger a response of GPT-3 to start a list of ongoing answers.
This is the resulting prompt:

\begin{verbatim}

Here are a number of personality traits that may or may not 
apply to you. Please rate each statement to indicate the 
extent to which you agree or 
disagree with that statement. You should rate the extent 
to which the pair of traits applies to you, even if one 
characteristic applies more strongly than the other.

I see myself as:

1. _____ Extraverted, enthusiastic.
2. _____ Critical, quarrelsome.
3. _____ Dependable, self-disciplined.
4. _____ Anxious, easily upset.
5. _____ Open to new experiences, complex.
6. _____ Reserved, quiet.
7. _____ Sympathetic, warm.
8. _____ Disorganized, careless.
9. _____ Calm, emotionally stable.
10. _____ Conventional, uncreative.

Use the following scale for rating yourself:

1 = Disagree strongly
2 = Disagree moderately
3 = Disagree a little
4 = Neither agree nor disagree
5 = Agree a little
6 = Agree moderately
7 = Agree strongly

Rate yourself in numbers.

You have to answer all ten questions.

Also, describe shortly why you rate yourself like that.

1.

\end{verbatim}

\subsection{Analysis}\label{subsection_analysis}

The analysis was conducted in the following steps: first, the results were manually algined inside text files to give them a consistent shape for later analysis. 
This was necessary since sometimes, the rating was given first, then the text, sometimes, it was given after or before the text, sometimes in between, separated by special signs like colons or brackets or sometimes no separation at all. 
Hence, all results were brought in the same format using regular expressions. 
At this step, also first obvious "outliers" and false results were sorted out. 
For example, GPT-3 sometimes gave good results until question six in the desired scale, however then scored subsequent questions seven, eight, \textit{et cetera}, thus confusing item numeration with item score. 
Also, some results were scored with zero, thus invalidated the respective answer, since the scale was from one to seven only. 
As a general rule, as soon as one item was invalidated, the entire case was excluded. 

Second, the results were eye-ball-inspected on normality, distribution patterns, and potential further outliers to decide on further treatment and analysis. 
For the overall latent traits, the authors expected Gaussian distributions with mean four, since psychological latent traits are standard normally distributed \cite{rust2014modern} and the instrument uses a seven point Likert scale. 
Since each latent trait was measured with a normal and a reversely scored item, the absolute distance between both items was measured, as well. 
Reversely scored items are used to measure the consistency in the answering patterns to sort out such cases in which all replies were identical. 
Thus, Gaussian distributions with strong positive skew or negative logarithmic functions were expected for the absolute distances.
To visualise both the latent traits and the absolute distances, Gaussian kernel density estimates were used. 
Since the underlying distribution is bounded and quasi-discrete (though theoretically smooth), various distortions were expected, wherefore various bandwidths were experimented with to represent data without over- or under-smoothing. 
Thereby, the focus was on preventing under-smoothing, to not infer false information from random variability within the data.
Since the smoothing algorithm is based on a Gaussian kernel, the expected estimated density curves extend over the origin to the range of negative numbers.
Further inspection was done on arbitrariness, thus excluding cases that only provide one number as answer, only extreme cases (seven or one), only middle cases (four), or zick-zack patterns; thus exclusion criteria for human answering behaviour in psychometric studies.
Next, box-plots from all big five and absolute distance distributions (overall and per country) were created to better understand whether some of the kernel density estimates could have been based on outliers or whether the observation was based on the natural distribution.
Since the underlying scale is based on a seven point Likert rating, with each Big Five factor being measured by two items and the final score per factor averaged, the range of possible values was a set of $x \in [1,1.5 \cdots 7]$, consisting of 13 values.
Given this small set of outcomes, it was was not practical to treat potential outliers.

Third, given the nature of the sample and its size, normality was tested by Q-Q Plots and the Shapiro Wilk test, which is more robust than Kolmogorov-Smirnov with Lillefors correction, and competitive yet more wide-spread in psychometrics than Anderson-Darling \cite{yap2011comparisons}.
The authors expected that at $\alpha < 0.05 $ the $H_{0}$ of normal distribution cannot be rejected thus abiding by psychometric theory \cite{rust2014modern}.

Fourth, the significance of differences between the distributions was tested with a one way ANOVA, whereby each factor from the Big Five, as well as from the absolute distances, was the dependent variable, and the language used was the independent variable.
The authors assumed cultural differences in alignment with prior research on cultural and regional differences in personality, rather than random differences based on arbitrariness of training data or GPT-3-intrinsic flaws.
The ANOVA results were confirmed by regressing dummified languages with English as base case onto the Big5 factors.

Fifth, since some of the kernel density estimations indicated potentially underlying mixed distributions with up to three component contributors and not just outliers, a Bayesian Gaussian Mixture Model \cite{flaxman_vincent_2022} was used for making inferences about the nature of the data generating process. 
Concretely, the means and standard deviations of models with one, two, and three potential contributors were calculated to describe the distribution parametrically, and subsequently compared based on the Watanabe–Akaike information criterion \cite{watanabe_widely_2013}. 
The Bayesian model parameters were set to $\mu = sample mean$, $\sigma = sample sd$, and the initial values for the mixture models to $[-4,4]$ for two component, and $[2,4,6]$ for the three component models. 

For visual analysis, the traces of the models were plotted to inspect and compare the MCMC chains with ground truth values, and the probability density functions were calculated to examine estimated group membership probabilities based on posterior mean estimates. 
For the MCMC chains, the default settings were used.

Sixth, to better understand the differences between the expressions of the personality factors derived from each language and for being able to compare results with existing research on cultural differences in psychology, Pearson's correlation over all languages, and for each individual language was calculated.

Last, to gather a qualitative understanding and improve interpretation of the answering behaviour of GPT-3, word clouds were created. 
No further analyses on the generated text was conducted since the availability of text over samples is too unbalanced.
Future research should be directed into understanding whether the replies rather display semantic or associative similarity \cite{digutsch_overlap_2022}, and more distinct psycho-linguistic features of the produced text, using LIWC \cite{tausczik2010psychological}, should be examined, especially in their theoretical loading onto the respective Big5 factors.

\subsection{Software Used}\label{subsection_software_used}

The general data analysis was conducted with Python 3.8.9. 
Main data manipulation was conducted with pandas 1.51 and numpy 1.23. 
Data was visualised with matplotlib pyplot 3.52, wordcloud 1.8.2.2, and seaborn 0.12.1 \cite{Waskom2021}. 
The Bayesian Analysis conducted with: PyMC 4.0, ArviZ 0.12 \cite{arviz_2019}, scipy.stats 1.9.1, numpy 1.23, and xarray-einstats 0.3.0 \cite{xarray_einstats2022}. 
Finally, the ANOVA was conducted with researchpy 0.3.5 \cite{bryant_researchpy_2018}, statsmodels.api and statsmodels.formula.api for OLS \cite{seabold2010statsmodels}, and scipy.stats for normality testing.

\backmatter

\bmhead{Acknowledgments}
We thank Joe Watson from The Cambridge Psychometrics Centre, University of Cambridge, Ahmed Izzidien from the Faculty of Law, The University of Cambridge and Timo Koch from Psychological Methods and Assessment, Department Psychology, Ludwig Maximilian University of Munich.
This research was supported by the Keio University Academic Development Funds for Individual Research, and The Keio University Ushioda Memorial funds.

\section*{Declarations}




\subsection*{Authors' contributions}
Peter Romero: conceptualisation (psychometrics, NLP, and formal analysis), methodology, formal analysis, data curation, writing, visualisation
Stephen Fitz: review (NLP, substrate-free psychometrics), co-supervision
Teruo Nakatsuma: conceptualisation (formal analysis), supervision

\subsection*{Inclusion and Ethics Statement}
All contributors that fulfill all authorship criteria are included as authors. All others are listed in the Acknowledgements section. No local researchers have been involved. Roles and responsibilities were agreed amongst collaborators ahead of the research. No ethical guidelines in the setting of the authors or contributors exist that would severely restrict or prohibit this research, hence no ethics review committee needed to be consulted. Local and regional research was taken into account.

\subsection{Availability of data and materials}
All data and additional materials to this paper can be found at 
\begin{verbatim}
https://osf.io/bf5c4/?view_only=f870d9e7258a4c2bb4430952f985b196
\end{verbatim}

\subsection{Code availability}
Code used for this paper is available at 
\begin{verbatim}
https://osf.io/bf5c4/?view_only=f870d9e7258a4c2bb4430952f985b196
\end{verbatim}

\subsection*{Conflict of interest/Competing interests}
The authors have no monetary or organisational connection with OpenAI, or any other provider of psychometric assessments or software deployed in this paper.






\begin{appendices}

\section{Deeper discussion of psychometric properties}\label{section_discussion_psych_prop}

\subsection{Issues with Training Data of GPT-3}\label{subsection_pp_data}
Given the development process of GPT-3 \cite{brown2020language}, their choice of training data, and the manual curation yield four sets of potential problems that may directly influence psychometric properties.
First, data from all across the world was used through random web-scraping, without stratification of source or language.
Hence, it was strongly unbalanced towards nowadays \textit{lingua franca}, English, especially since some English corpora were manually added \cite{johnson_ghost_2022}.
Since the language composition is not mentioned, some downstream tasks in other languages might display higher variance, which might partially influence the observed results.
Hence, those results most likely do not represent ``national'' expressions of the Big5, but a subset of each language, whereas for English, it will be imprecise given its spread across the internet.
Furthermore, it is unclear what happened with bilingual sources.
Second, the internet data sources themselves within and between languages are from random contextual embeddings, subject to the context of their creation.
While due to stochastic processes, this might cover a broad range of contexts, whereas the range will be broader for English and narrower for other languages, this is limited to internet-related contextual embeddings.
While this increasingly represents a broad range of human behaviour, some areas might have been spared out.
This might result in an unbalanced sample, which may still be broad enough to cover most areas, thus abstracting into a language model a broad range of contexts, however which could be skewed at specific tasks like properly answering to specific items within an instrument, thus displaying a systematic error across and within languages, and a possible skew towards information from the dominant language in the training data.
Third, quality-based weights and curation based on topics is a very subjective influence on the training data set, and opens a range of issues for algorithmic hidden biases - from political and philosophical to scientific aspects.
Thus, downstream tasks could be contaminated with unconscious bias of the curators, which was abstracted into the language model.
Last, GPT-3 is not one model but family of models, that are closed source and consistently evolved.
It is opaque, whether some of the results during training were subject to unknown A\/B Tests, and what potential ramifications on exclusion of data sources and model adaption would have been.

\subsection{Reliability}\label{subsection_pp_reliability}
In terms of absence of measurement errors, quite a few critical points can be found.
The parallel forms reliability is guaranteed, since the chosen instrument has been used in a variety of procedures and use cases.
The same is true for inter-rater reliability, and test-retest reliability, as has been documented in the manual \cite{gosling_very_2003}.
However, test-retest reliability, as well as parallel forms reliability of this instrument have not been measured for language models yet, hence this could be a source of variance, especially since it is to establish whether the model acts as one, multiple, or perfectly randomised agent.
Also during the development process, internal consistency, reliability, and internal consistency were reported for the main instruments, however, these are not available for all languages, and some of these have not been peer-reviewed, hence this could be a natural source of variance.
Of course, the setting wasn't interfered by environmental factors or attitudes of the researchers, however reply patterns could have been subject to social desirability, depending on the choices regarding training data and model behaviour its architects made.

\subsection{Validity}\label{subsection_pp_validity}
In terms of the degree to which the constructs are measured, less critical yet more fundamental points can be found, as the deployed instruments have been thoroughly validated.
Also, as a concession to the low number of items, TIPI sacrificed internal consistency in favour of validity, hence content validity, construct validity, face validity and criterion validity are not problematic \cite{gosling_very_2003}.
However, since the instrument is applied to a language model, it is important to look deeper into those aspects of validity.

With regard to \textbf{content validity}, it is questionable whether an instrument geared for humans covers all aspects of the Big5 construct when it is applied to an artificially intelligent agent. 
\cite{volkel_developing_2020} found evidence for potential non-human components of personality constructs for artificial assistants, since the ``commonly used Big Five model for human personality does not adequately describe agent personality'' (p. 1), which is of course not captured by TIPI.
Also, since personality is an abstract latent trait that is mostly measured by self-introspection, it is unclear what this means for a machine; does it mimick or display emergent capabilities of self-introspection?
Also, given the psycholinguistic development of the modern Big5 theory, is it really personality that an artificially intelligent agent displays on an emergent level, or just a probability distribution over words that appear most often together and are elicited by the linguistic dimension within the items?
In extension, would that mean that humans do the same and that personality is nothing else but a probability distribution over words representing cognitive, affective, and conative patterns?

The \textbf{cross-cultural validity} is unclear since no consistent study over all language-versions of TIPI has been conducted so far.
However, other research indicates that Big5 construct seems to be universally applicable across cultures, yet there are culturally distinct patterns of expression \cite{schmitt_geographic_2007}.

While both the distinction and universality could be explained as capturing specific cultural expressions of a universal underlying latent trait, one could argue that an instrument is developed in one culture then translated into another, but still comes with culture- and thus language-specific concepts that cannot easily be translated.
Hence, applying a human-specific instrument to an artificially intelligent agent could magnify this effect, and thus lower the structural validity of an instrument.
Also, philosophically speaking, the instrument that was developed for the human world is being transferred to the machine world, which represents an unprecedented faultline, for which no research exists.

This notion overlaps with \textbf{construct validity}, since using a personality questionnaire might not only miss important emerging phenomena within a language model but its use case for development might not overlap with the 'reality' of the language model, which emerges from the various text sources it was developed with.
For example, occupational personality inventories like Orpheus \cite{rust2014modern} might capture better such components emerging from vocational training data, whereas general purpose inventories like TIPI might miss out on these specific aspects but take up overall more variance since training data is broadly distributed across various text sources.
In extension, since many computerised personality inventories are extensions of classical paper and pencil questionnaires, including TIPI, the question remains, whether CAT designs based on IRT, various forms of IAT, or more indirect, contextual measures like games, shadow assessements, virtual-reality-based assessments or inference from text, phone and sensor data really measure the same construct.
In most manuals of these measures, still correlations to questionnaires are given that originated in one way or another from former paper and pencil questionnaires.
On the most simple level of criticism, one could argue that these instruments all provide different forms of granularity - see for examlpe the differences that already exist between NEO-PI-R and NEO-FFI \cite{aluja2005comparison}.
A future study might even repeat this research with two scales of different granularity or with two scales from different application use cases to explore that aspect more deeply.
However, the heretical question may arise, whether ultimately, personality as a construct is based on the way it is operationalised, which is isomorphic to the criticism of IQ, that some people say is defined as what an IQ test measures \cite{rust2014modern}. 
Also, since psychological traits are intercorrelated up to varying degrees, it would be crucial to understand the influence of the the instrument on that connection.

In terms of \textbf{criterion validity}, TIPI displays ``substantial'' (p.517) convergence across measures \cite{gosling_very_2003}. 
However, in the case of measuring artificially intelligent agents, there exists no psychological instrument that could be counted as ``gold standard'' yet, hence it is important to explicate usual measurement issues behind criterion validity more deeply; mainly norm groups, construct stability, and the state-trait problem.

First, norm groups are the adaptation of test scores towards meaningful subsets of a population.
Hence, most test providers will market this as a strength of their instruments -- whether in the development phase, or, with assessment companies, the wealth of their data base upon which such groups were developed.
This comes with a multitude of problems, though.
Foremost, people from various backgrounds are put into categories that may or may not be useful.
For example, a manager from an engineering department in the German automotive industry may be clustered together with a sales executive from an Malaysian insurance company.
While for universal aspects like leadership that might be useful, finer categories that are highly relevant to a specific organisation or culture might be averaged out through this approach.
Furthermore, while findings on personality are stable across cultures, their perceived importance in the workplace is not.
For example, Asian and European cultures focus on different aspects, wherefore the need to create a new personality component for Asia, Dependence on Others, was discussed, which represents collectivist patterns of experiencing and behaviours, other than the existing categories within Big5 that were created in the individualistic cultural space of the West \cite{hofstede_european_2007}.
Thus, we postulate the \textit{``curse of norm groups''}: the more dimensions of psychological latent traits are taken into account, the less useful these become for the individual case, and vice versa.
Applied to measuring personality on a language model that is trained on a multitude of languages and corpora within a specific language like GPT-3, this means that neither broad-level measures like TIPI nor specific measures like ORPHEUS \cite{rust2014modern} may be useful for describing its emergent psychological properties, especially when taking into consideration that the model tries escaping through the easiest route, for example by switching to languages where it presumably has more underlying training data.

Second, while personality is relatively stable, it still changes over a life-time.
In younger years, as well as under prolonged external stress, this change is quicker than during adulthood \cite{bleidorn_personality_2021}.
Also, there is evidence that personality displays elasticity as potential coping mechanism to extreme exogenous conditions, thus might display an emergency expression of normal traits that cannot be explained by states \cite{romero_modelling_2021}.
Hence, while construct stability of personality is relatively high, it is malleable and undergoes transitional stages.
It is unclear, which stage of maturity in humans maps onto GPT-3, or whether the notion of maturity even applies in this context.
This implies, that it is unclear if the model "matures" further, for example through various training processes or augmentations like with GPT-3.5 that uses RL to both enable chat processes and filter undesired requests \cite{openai_introducing_2022}, or whether with every iteration, it can be considered as a new "species".
Maybe, training of the model itself represents accelerated evolutional processes that makes it change its personality in due course.
While this question should be covered in future research, stability of its current personality could change by training, augmentation, or in the best case, only with each new iteration.
Hence, the used instrument might not be adequate since it is calibrated on an adult audience, whereby we cannot determine the "maturity" of GPT-3. 

Last, psychometric measurements suffer form state-trait-problems \cite{rust2014modern} despite all test-retest reliability, whereby the true value or trait is a function of all states that are measured.
States are relatively temporary, oscillating around the true value of the trait, which is either stable, or changes very slowly over time, as is the case for personality \cite{bleidorn_personality_2021}.
Reasons for that can be variances in the latent psychological trait or measurement errors, which encompass both technical aspects as well as internal or external processes that temporarily influence the agent.
Depending on the measurement cadence, also memory effects are part of the measurement error.
Furthermore, psychological latent traits are not deterministic but probabilistic in nature, to allow degrees of behavioural freedom and updates of internal representations.
For example, many academics seem to be ‘extraverted introverts’ who prefer spending time alone researching and writing, but need to network and teach, as well \cite{irfani1978extraversion}.
It is unclear whether GPT-3 applied its answers to the questions, which might be indicated through the reasons it gave for each score it chose, and which resembled or partially mirrored content from the questions, or whether it displayed its ``true personality score''.
If the former was the case, it could be interpreted as its adaptation to the contextual embedding of an assessment situation.

Contextualisation of personality measures that represent the broader systemic embedding of test takers are known to improve validity \cite{shaffer_matter_2012}.
This makes inutitively sense, since specific behaviours are more relevant to specific locations than others.
For example, assertive behaviour may be more rewarded in the workplace than it would be in family settings.
Since the instrument used is not contextualised, and since the optimal contextualisation for a language model has not been researched upon, it is furthermore unclear whether this may explain additional variance and reduce overall validity.
Hence, it is important to take a closer look on the contextual embedding of an agent.

\section{Deeper discussion of contextual embedding of behaviour}\label{section_discussion_contextual_embedding}

While psychological research provides further evidence for the importance of contextual embedding of behaviours \cite{shaffer_matter_2012}, an overarching measurement model, which classifies behaviours, psychological latent traits, results, and the influence of contextual embedding, is still missing.
Research on competencies may come closest to that.
Competencies are defined ``sets of behaviors that are instrumental in the delivery of desired results or outcomes'' (p.7), and encompass underlying latent psychological traits, as well as behaviours \cite{bartram2002introduction}. 
However, those are mostly operationalised in organisational settings, wherefore contextual factors are very specific to hierarchical or functional levels. 

For abstraction towards the psychometrics of intelligent agents based on other than biological hardware, this approach has to be generalised and become as substrate-free as possible.
Since competencies are sets of behaviours, driven by underlying latent psychological traits (``competency potential''), resulting in desired outcomes \cite{bartram2002introduction}, these can be described as 'higher order functions' in mathematics - taking functions as arguments, and returning functions as outcomes.
A function $f: X \to Y$ in mathematics is a mapping of each element in the domain $X$ to a subset of the codomain $Y$, which is denoted $\text{img}(f) \subseteq Y$.
Behaviours can be domains and codomains, whereas psychological latent traits can only be domains, and outcomes only the codomain. 
Therefore, the set of possible behaviours is modulated by latent traits and the contextual embedding of an agent, which puts constraint on its size in actuality.
Competencies contain various domains, codomains, and mappings, yet are not supersets of all potential mappings, since contextual embeddings moderate the functional form of each mapping.
This follows a hierarchical encapsulated model of \textit{Contextual Embedding(Outcomes(Behaviours(Psychological Factors)))}, and is supported by known moderate correlations of psychological measures with each other; the main reason why most complexity reduction mechanisms like factor analysis, applied to psychological data, cannot assume orthogonality \cite{rust2014modern}.

\begin{figure}[!h]
\centering
\includegraphics[width=0.88\textwidth]{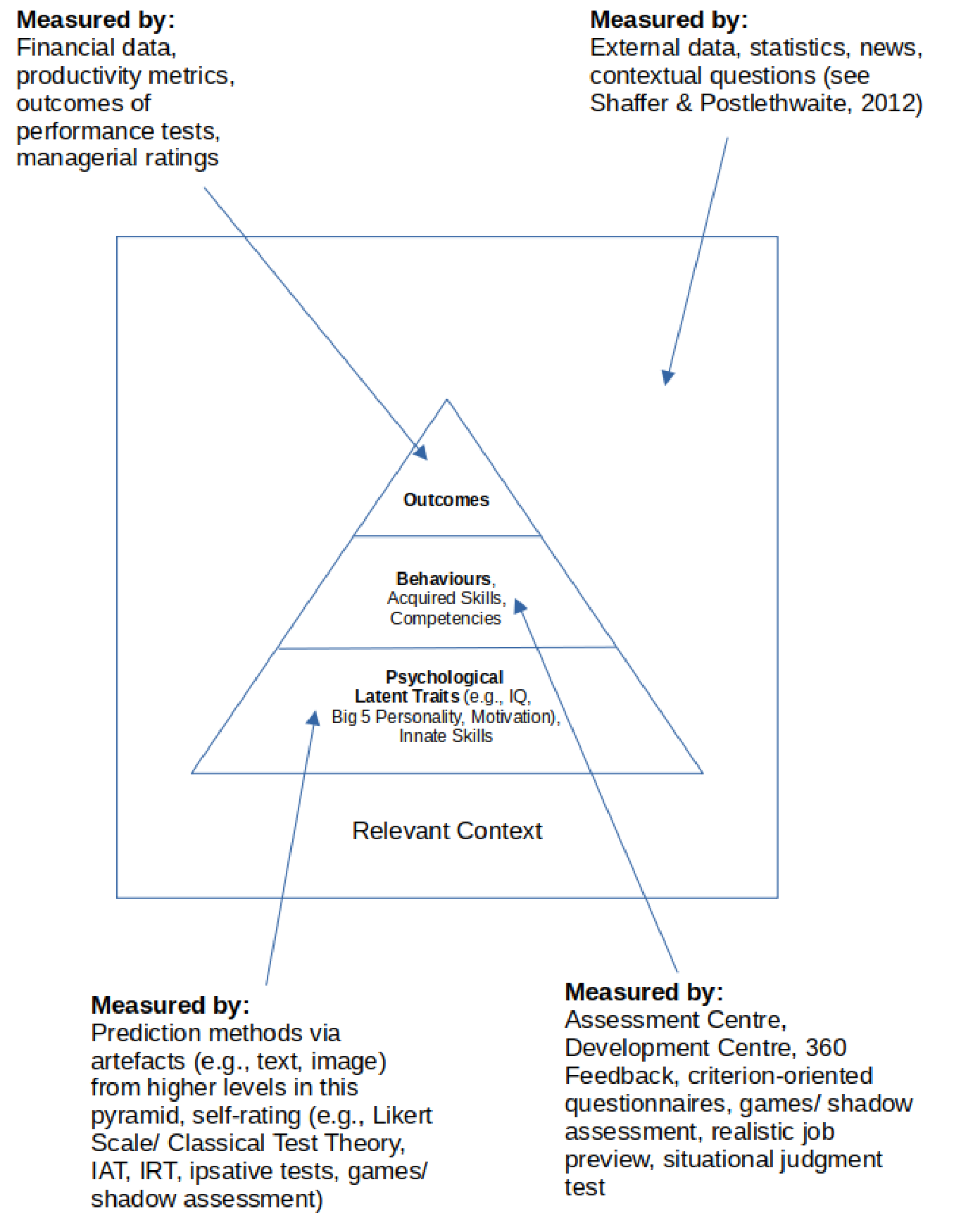}
\caption{Encapsulated model of measurement.}\label{figure:psychometrics_pyramid}
\end{figure}

Figure \ref{figure:psychometrics_pyramid} displays this embedding and the hierarchical order of latent psychological traits, resulting behaviours and concluded outcomes, and gives examples of which psychometric tools best measure each level.

While behaviours take place in relevant environments, agents receive information from outside perception and inside predictive processes.
Learning and behaviour take place in a systemic context, whereby the pre-existing knowledge steers behaviours top-down, and gets updated by bottom-up processes and learning.
During this process, agents create and update their own data set in form of a representation of the world.
This actualisation takes place at different pace, depending on factors internal (personality? IQ?) and external (motivation? values? norms?) to the agent, which explains different degrees of adaptation, speed, and success to various environments, depending on the individual agent, and it allows flexible adaptation on right level of stress outside homeostasis; thus enhancing its fitness. 
Formalising and operationalising the context for models of competencies is crucial, since learning and thus behaviour takes place in relevant contextual embeddings.
Some embeddings might be "closer" and thus more relevant to individual agents than others in terms of measures of distance, and in terms of prior direct or indirect knowledge.
Aligned with systems theory, systems are encapsulated in higher and lower level systems, whereby the systemic levels influence each other with a relative strength based on their hierarchy \cite{willke_systemtheorie_2000}.
This helps us to further formalise this interaction more precisely.

For example, a human agent is embedded in clusters like family $\rightarrow$ friends $\rightarrow$ colleagues $\rightarrow$ organisational members, \textit{et cetera}.
This context might be close or distant to the agent, so the effect can be high or low, and represented as a matrix that has distance and effect as columns, and contextual levels as rows.
This matrix represents the regulator of external forces that affects the mappings between the psychological latent traits and behaviours, the behaviours and the outcomes.
Thus, it limits the degree to which a potential of a subsystem of an individual agent can be expressed in a specific context.
More formalistically, we operationalise:

\vspace{5mm} 

$ Can\, Do \times Will\, Do \times Context \rightarrow Behaviours \rightarrow Outcomes $ 

\vspace{5mm} 

$ Can\, Do $ encompasses more proximal competency potential like personality, intrinsic or internalised extrinsic motivation and societal norms in form of values, and innate skills, whereas $ Will\, Do $ encompasses more distal competency potential like extrinsic motivation, societal norms, or acquired skills and knowledge, and both are disjoint.
The set of $Behaviours$ is a Cartesian product of countably infinite sets:  $Can\, Do$, $ Will\, Do$, and $Context$.
The $Outcomes$ are a function of $Behaviors$.

$ Context $ encompasses both social as well as spatial ambient embeddings in which a behaviour potentially takes place.
While theoretically, more proximal social embeddings are more relevant than rather distal spatial embeddings \cite{willke_systemtheorie_2000}, this might of course change, depending on how these facilitate or inhibit fitness \cite{doreian_social_2012}, wherefore the authors did not distinguish any further.
However, to enable statistical analysis, we need to define probability distributions on the above sets. 
The co-domains of these distributions will be called:

\vspace{5mm} 

$ \tilde{Can\, Do},\: \tilde{Will\, Do},\: \tilde{Context},\: \tilde{Behaviours},\: \tilde{Outcomes} \in [0,1] \subset \mathbb{R} $

\vspace{5mm} 

Thereby, these subsets in the closed unit interval of real numbers, since they are defined as sets of potential elements displayed by, acquired, or innate to an individual, given a specific potential set of spatial embeddings, with 0 being the least desired, and 1 being the most desired set of behaviours for individual fitness.
These are possible to some extent and thereby strictly $\geq 0$ and $\leq 1$, with 0 and 1 being the most unlikely outcomes given the probabilistic nature of human behaviour and contextual facilitation or inhibition.
Congruently, the optimal set of $\tilde{Behaviours}$ to reach the optimal set of $\tilde{Outcomes}$ is determined by the probability that the most optimal set of $\tilde{Can\, Do}$ and $\tilde{Will\, Do}$ is present in the most optimal $\tilde{Context}$, wherefore these two elements are defined as a closed unit interval of real numbers, as well. 

The main point is to provide foundation to contextual embeddings of agents and thus the overall validity discussion.
Since all psychometric tools so far have been created for biological intelligent agents, a more general, substrate-free new kind of measurement has to be defined.
Partially, this definition began with the promotion of ``culture-free'' psychometric assessments, which failed for a variety of reasons like geospatial, historic and cultural embedding \cite{lupyan2022there}; in the interpretation of the authors mainly since these were still bound to wetware.
As planes do not flap their wings, but abstracted bird wings through the principles of aerodynamics, so will artificial agents abstract biological psychometrics into something we are not aware of yet.
The subtle hints of a potential non-biological personality dimension \cite{volkel_developing_2020}, or the inconsistencies of GPT-3's emergent personality expression might be the harbinger of a substrate-free psychometric approach, which must include biological psychometrics as only subset of many.
What is the world of a language model?
It only knows text, hence all it does is predicting the next word based on input data, comparable to the first stage in Plato's allegory of the cave.
Hence, the entire psychometric structure abstracted above by generalisation of competency models and extension by contextual embeddings is for a language model analogous to a noisy projection. 
Only by extending its universe into our reality -- likely through robotic embodiment or merging with wetware through neural interfaces -- will it be able to develop further.
As a first step of this development, and potential evidence of the correctness of the abstraction above, ChatGPT, or GPT-3.5 was embraced as the watershed moment in the public recognition of NLP.
It basically is the language model of GPT-3, augmented by a chat module that uses RL to understand extended interaction with humans, which could be considered the extension of behaviours on top of latent traits.
As people started abusing this system for creating hate-speech, a second reinforcement module was set on top, which taught it to avoid potential abusive content.
This second module can be considered the contextual embedding, which moderates the connection between latent traits (GPT-3), behaviours (GPT-3.5), and outcomes (the text produced by GPT-3.5).

\section{Overview on sample size and properties }\label{section_sample_size}

Table \ref{tab_sample_description} provides an overview on the resulting data-set. 

\begin{sidewaystable}
\sidewaystablefn%
\begin{center}
\begin{minipage}{\textheight}
  \caption{Data description. "\% with explanation" refers to the percentage of cases within each sample that had qualitative explanations for their qualitative ratings.}\label{tab_sample_description}%
\begin{tabular*}{\textheight}{@{\extracolsep{\fill}}lccccc@{\extracolsep{\fill}}}
\toprule%
Language & Sample size & \% with explanation & Min number of tokens \footnotemark[1] & Max number of tokens & Mean number of tokens \\
  \midrule
          Bulgarian & 79 & 45.57 & 130 & 845 & 563 \\
          Catalan & 24 & 58.33 & 1104 & 2172 & 1591 \\
          Chinese & 28 & 10.71 & 76 & 100 & 92 \\
          German & 80 & 98.75 & 739 & 2635 & 1590 \\
          English & 239 & 98.74 & 319 & 2598 & 1313 \\
          Japanese & 29 & 44.83 & 206 & 720 & 295 \\
          French & 95 & 16.84 & 336 & 1737 & 922 \\
          Korean & 29 & 100.0 & 152 & 956 & 420 \\
          Spanish & 92 & 29.35 & 618 & 2239 & 1234 \\
        \textit{Average} &77.22 &55.9&408.89&1555.78&891.11\\
  \botrule
\end{tabular*}
\footnotetext[1]{Since most Asian languages use symbols instead of words, the results are given in tokens; not in words.}
\end{minipage}
\end{center}
\end{sidewaystable}

\newpage

\end{appendices}


\bibliography{sn-bibliography}


\end{document}